\documentclass[letterpaper]{article} % DO NOT CHANGE THIS
\usepackage{aaai2026}  % DO NOT CHANGE THIS
\usepackage{times}  % DO NOT CHANGE THIS
\usepackage{helvet}  % DO NOT CHANGE THIS
\usepackage{courier}  % DO NOT CHANGE THIS
\usepackage[hyphens]{url}  % DO NOT CHANGE THIS
\usepackage{graphicx} % DO NOT CHANGE THIS
\usepackage{natbib}  % DO NOT CHANGE THIS AND DO NOT ADD ANY OPTIONS TO IT
\usepackage{caption} % DO NOT CHANGE THIS AND DO NOT ADD ANY OPTIONS TO IT
\usepackage[utf8]{inputenc} % allow utf-8 input
\usepackage[T1]{fontenc}    % use 8-bit T1 fonts
\usepackage{url}            % simple URL typesetting
\usepackage{booktabs}       % professional-quality tables
\usepackage{amsfonts}       % blackboard math symbols
\usepackage{nicefrac}       % compact symbols for 1/2, etc.
\usepackage{microtype}      % microtypography
\usepackage{xcolor}         % colors
\usepackage{graphicx} 
\usepackage{amsmath} 
\usepackage{float}  % Add in your preamble
\usepackage{mathtools}

\title{Towards a Foundation Model for Partial Differential Equations Across Physics Domains}

\author{
    Eduardo Soares\textsuperscript{\rm 1},
    Emilio Vital Brazil\textsuperscript{\rm 2},
    Victor Shirasuna\textsuperscript{\rm 1},
    Breno W. S. R. de Carvalho\textsuperscript{\rm 2},
    Cristiano Malossi\textsuperscript{\rm 3}
}
\affiliations{
    \textsuperscript{\rm 1}IBM Research Brazil, Sao Paulo, Brazil\\
    \textsuperscript{\rm 2}IBM Research Brazil, Rio de Janeiro, Brazil \\
    \textsuperscript{\rm 3}IBM Research Zurich, Rüschlikon, Switzerland\\
    eduardo.soares@ibm.com, evital@br.ibm.com, vshirasuna@ibm.com, brenow@ibm.com,acm@zurich.ibm.com
}

\begin{document}

\maketitle

\begin{abstract}
We present \textbf{PDE-FM}, a modular foundation model for physics-informed machine learning that unifies spatial, spectral, and temporal reasoning across heterogeneous partial differential equation (PDE) systems.  
PDE-FM combines spatial–spectral tokenization, physics-aware conditioning, and a Mamba-based state-space backbone with an operator-theoretic decoder, enabling scalable and data-efficient modeling of complex physical dynamics.  
In contrast to task-specific neural operators, PDE-FM is pretrained once on diverse PDE datasets and can be transferred to new physical regimes without architectural or data-specific modifications.  
Evaluated on twelve 2D and 3D datasets from \textit{The Well} benchmark—spanning hydrodynamic, radiative, elastic, and astrophysical phenomena—PDE-FM achieves state-of-the-art accuracy in six domains, reducing mean VRMSE by 46\% relative to prior operator-learning baselines.  
The model demonstrates robust \textit{cross-physics generalization}, excelling in turbulent and radiative systems while maintaining strong performance in linear and steady-state regimes.  
These results suggest that large-scale pretraining across diverse physical processes can yield transferable representations of dynamics, marking a step toward unified, foundation-level surrogates for multi-physics simulation and scientific discovery.
\end{abstract}

\section{Introduction}

Over the past decade, \textit{neural operators} and \textit{physics-informed learning} have reshaped how we approximate and reason about complex spatiotemporal systems \cite{raissi2017physics, goswami2023physics}.  
These approaches replace traditional numerical solvers with data-driven surrogates that learn mappings between functional spaces, enabling efficient simulation and prediction in high-dimensional physical systems.  
Architectures such as the Fourier Neural Operator (FNO) \cite{li2020fourier, li2023fourier}, Transformer-based operator networks \cite{wang2024advancing, hao2023gnot}, and U-net-style surrogates \cite{comlekoglu2025surrogate, shen2025vortexnet} have demonstrated remarkable ability to capture intricate solution manifolds of partial differential equations (PDEs).  
However, most existing operator-learning frameworks remain \textit{domain-specific}—trained on isolated datasets, fine-tuned to narrow classes of PDEs, and constrained by inductive biases that limit transfer across physical regimes \cite{alesiani2022hyperfno, wang2022generalizing, hu2021extended}.  
As a result, each model functions as a bespoke surrogate, effective only within the regime it was trained for, with performance rapidly degrading when boundary conditions, scales, or governing dynamics change \cite{krishnapriyan2021characterizing, shi2024physics}.

This fragmentation stands in contrast to recent trends in machine learning toward \textit{foundation models}—large, pre-trained architectures that integrate information across diverse domains to yield transferable representations \cite{bommasani2021opportunities, wang2023large}.  
In natural language and vision, such models have transformed generalization and data efficiency, yet the extension of this paradigm to scientific modeling remains in its infancy \cite{touvron2023llama, zhai2022scaling}.  
Physical systems pose unique challenges: data are multi-resolution and multi-scale \cite{pathak2022fourcastnet, yang2025multi}, governed by constraints such as conservation laws \cite{karniadakis2021physics}, symmetry, and stiffness; they evolve in continuous space-time \cite{raissi2019physics, angelov2010evolving}; and they couple nonlinear operators across disparate physical processes \cite{aarts2025physics, li2025multi, sun2024physics}.  
A foundation model for physics must therefore reconcile two seemingly opposed requirements: (1) the scalability and generalization of large sequence models \cite{wiesner2025towards, alkin2024universal}, and (2) the physical fidelity and inductive structure of domain-specific solvers \cite{chalapathi2024scaling, gao2025can}.

\textbf{PDE-FM}, \textit{Partial Differential Equation Foundation Model}, is introduced to address this gap.  
PDE-FM is a modular architecture that combines \textit{spatial and spectral tokenization}, \textit{physics-aware conditioning}, and a \textit{Mamba-based state-space backbone} \cite{gu2023mamba} with an \textit{operator-inspired decoder} \cite{tiwari2025latent}.  
This hybrid design bridges symbolic physics priors and data-driven scalability: spatial and spectral tokenization encode multi-resolution field structure; physics-aware embeddings enforce consistency with PDE invariants; and the Mamba backbone captures long-range temporal and spatial dependencies in linear time.  
Together, these components enable PDE-FM to serve as a \textit{general-purpose surrogate}—a pretrain-once, adapt-everywhere framework for multi-physics simulation.

\begin{figure*}[t]
    \centering
    \includegraphics[width=1\linewidth]{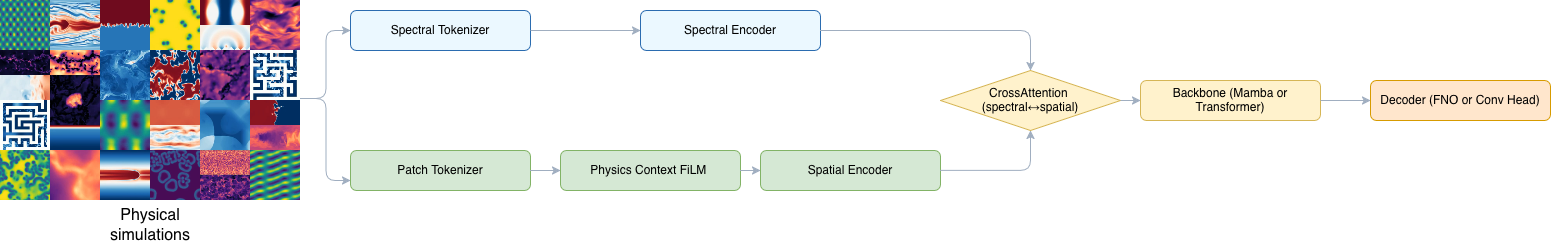}
    \caption{General architecture of PDE-FM.}
    \label{fig:fmpde}
\end{figure*}

We pretrain PDE-FM across twelve heterogeneous 2D and 3D datasets from \textit{The Well} benchmark suite \cite{ohana2024well}, spanning hydrodynamic, radiative, elastic, and astrophysical phenomena.  
This corpus includes regimes ranging from low-Reynolds active suspensions and radiatively cooled multiphase flows to elastic turbulence and relativistic magnetohydrodynamics, providing a diverse substrate for cross-physics representation learning.  
Our experiments show that PDE-FM achieves state-of-the-art accuracy in six datasets and ranks second in five others, with a mean VRMSE reduction of over 40\% relative to prior neural operator baselines.  
The model generalizes robustly across nonlinear and turbulent domains, such as Rayleigh–Bénard convection, shear flow, and radiative turbulence, while maintaining competitive accuracy in steady or linear regimes like Helmholtz scattering.  
These results suggest that large-scale pretraining over diverse physics regimes induces emergent \textit{cross-physics generalization}, where representations learned from one physical family transfer beneficially to others.

Beyond empirical performance, PDE-FM illustrates a new design space for scientific machine learning: scalable models that \textit{learn operators as distributions over physics}, rather than as isolated mappings.  
By combining operator-theoretic structure, spectral reasoning, and state-space recurrence within a unified framework, PDE-FM bridges the conceptual gap between neural operators and foundation models.  
We view this as a step toward a broader class of multi-domain scientific foundation models capable of learning transferable physical representations across scales, geometries, and governing equations.

\section{Methodology}

PDE-FM is a modular foundation model that learns solution maps from heterogeneous physical simulation datasets. 
Given input fields \(u \in \mathbb{R}^{C\times H\times W}\), where \(C\), \(H\), and \(W\) denote the number of channels, height, and width of the spatial domain respectively, 
we (i) tokenize spatial patches and low-frequency spectra, (ii) fuse modalities via cross-attention under physics-aware FiLM conditioning, 
(iii) model long-range dependencies with a Mamba state-space backbone, and (iv) decode with a shallow Fourier operator layer. 
Training employs a dual spatial–spectral objective and a multi-dataset curriculum with dataset-specific adapters (Figure~\ref{fig:fmpde}).

Let \(d\) be the token embedding dimension, \(p\) the physics context dimension, \(m\) the spectral truncation (modes per axis), \(h\) the number of attention heads, and \(N_p=(H/p_s)\times(W/p_s)\) the number of patches for patch size \(p_s\). Unless noted, tensors are batch-first.

\subsection{Tokenization and Physics-Aware Conditioning}
We build a dual representation
\begin{align*}
  T_{\text{spatial}} &= \mathrm{PatchConv}(u)\in \mathbb{R}^{N_p\times d}, \\
  T_{\text{spectral}} &= \mathrm{Linear}\!\big(\mathrm{FFT}_m(u)\big)\in \mathbb{R}^{1\times d},
\end{align*}
where \(\mathrm{FFT}_m\) keeps the lowest \(m\times(m/2{+}1)\) frequencies per channel (real/imag stacked). 
To incorporate physics metadata \(c\in\mathbb{R}^{p}\) (e.g., boundary conditions, constitutive parameters, time grids), we apply FiLM modulation~\cite{perez2018film} to spatial tokens:
\begin{equation*}
  \tilde{T}_{\text{spatial}} = T_{\text{spatial}}\odot\big(1+\gamma(c)\big)+\beta(c),\quad 
  \gamma,\beta:\mathbb{R}^{p}\!\to\!\mathbb{R}^{d}. 
\end{equation*}
We prepend a learned context token \([\mathrm{CLS}]\) when \(c\) is present. 
Patches capture locality and boundary effects; a global spectral token carries coarse global structure and smoothness priors; FiLM enables explicit parameter control.

FFTs run in FP32 for numerical stability; missing \(c\) defaults to zero vectors. We use dataset-level standardization for \(c\).

\subsection{Dual Encoders and Cross-Modal Fusion}
Spatial tokens pass through ConvNeXt-style residual blocks~\cite{liu2022convnet}; while spectral tokens pass through an MLP:
\begin{align*}
\hat{T}_{\text{spatial}} &=\mathrm{SpatialEnc}(\tilde{T}_{\text{spatial}}),\\
\hat{T}_{\text{spectral}} &=\mathrm{SpectralEnc}(T_{\text{spectral}}).
\end{align*}

We perform shallow bidirectional cross-attention:
\begin{align*}
  \hat{T}_{\text{spatial}} &\leftarrow \mathrm{Attn}(\hat{T}_{\text{spatial}}, \hat{T}_{\text{spectral}}),\\
  \hat{T}_{\text{spectral}} &\leftarrow \mathrm{Attn}(\hat{T}_{\text{spectral}}, \hat{T}_{\text{spatial}}),
\end{align*}
with \(\mathrm{Attn}(Q,K,V)=\mathrm{softmax}(QK^\top/\sqrt{d/h})V\).
A single spectral token gates global context into spatial tokens without quadratic cost. 

\subsection{Long-Context Backbone}
We concatenate \([\mathrm{CLS}]\), spatial, and spectral tokens to obtain \(T\in\mathbb{R}^{(N_p+1)\times d}\) and process with a Mamba state-space model~\cite{gu2023mamba}:
\[
  T^{(l+1)} = T^{(l)} + \mathrm{MambaLayer}\big(T^{(l)}\big),\quad l=1,\dots,L.
\]
Mamba provides sub-quadratic \(\mathcal{O}(N_p d)\) compute and memory vs.\ \(\mathcal{O}(N_p^2)\) attention, enabling large grids and long contexts.  We stabilize training via layer normalization before the backbone and gradient clipping.

\subsection{Spectral Operator Decoder}
We reshape the spatial slice back to a latent grid \(z\in\mathbb{R}^{d\times H/p_s\times W/p_s}\), upsample to \((H,W)\), and decode with a shallow 2D FNO~\cite{li2020fourier}:
\begin{equation*}
  \hat{u}(x) = \sum_{|k_x|\le m_x,\,|k_y|\le m_y} W_{k}\cdot \mathcal{F}[z](k)\,e^{2\pi i k\cdot x}.
\end{equation*}
The FNO head biases toward spectral smoothness while keeping capacity in the backbone.

We minimize a dual spatial–spectral objective

\begin{equation*}
\begin{aligned}
\mathcal{L} &=
\underbrace{\sqrt{\tfrac{1}{|\Omega|}\sum_{x\in\Omega}\big((\hat u(x)-u(x))-\mu\big)^2}}_{\mathclap{\text{VRMSE}}}
\\
&\quad + \lambda\,
\underbrace{\tfrac{1}{|\mathcal{K}|}\sum_{k\in\mathcal{K}} w(k)\,\lVert \hat U(k)-U(k)\rVert_2^2}_{\mathclap{\text{Spectral L2}}}\, .
\end{aligned}
\end{equation*}

where \(\mu=\frac{1}{|\Omega|}\sum_{x}(\hat{u}(x)-u(x))\), \(\mathcal{K}\) is the truncated frequency set, and \(w(k)\) increases with \(\|k\|\) to emphasize high frequencies.

When invariants are available, we add
\[
 \mathcal{L}_{\text{cons}} = \sum_{j}\alpha_j\,\big|\mathcal{I}_j(\hat{u})-\mathcal{I}_j(u)\big|
 \quad\text{and}\quad
 \mathcal{L}_{\text{PDE}}=\beta\,\|\mathcal{R}(\hat{u})\|,
\]
for conserved quantities \(\mathcal{I}_j\) (e.g., mass, energy) and residual \(\mathcal{R}\) of the governing PDE.
We cosine-anneal \(\lambda\) and (if used) \(\alpha_j,\beta\), starting with higher spectral weight to warm-start global structure.

\subsection{Multi-Dataset Pretraining}
We consider datasets \(\{\mathcal{D}_i\}\) from \textit{The Well} with heterogeneous channels \(C_i\).
Dataset-specific \(1{\times}1\) adapters normalize into a shared latent channel budget \(L\):
\begin{equation}
  x_i^{\text{lat}} = A_i^{\text{in}}(x_i) \in \mathbb{R}^{(L\cdot h_{\!i})\times H\times W},\quad
  \hat{y}_i = A_i^{\text{out}}\big(f_\theta(x_i^{\text{lat}}, c_i)\big),
\end{equation}
where \(h_i\) is the history length for \(\mathcal{D}_i\) and \(f_\theta\) is the shared core.

Batches are drawn with probability
\[
  p(i) \propto \big(\varepsilon + \overline{\mathcal{L}}_i\big)^\alpha \cdot |\mathcal{D}_i|^\tau,
\]
combining temperature scaling (\(\tau\)) with difficulty-aware weighting via the EMA loss \(\overline{\mathcal{L}}_i\) (exponent \(\alpha\in[0,1]\)). 
This reduces overfitting to large/easy datasets and mitigates negative transfer.

Tokenization is \(\mathcal{O}(N_p d + C m^2)\); fusion is \(\mathcal{O}(N_p d)\); Mamba is \(\mathcal{O}(N_p d L)\); the FNO head is dominated by truncated FFTs \(\mathcal{O}(C H W \log(HW))\) with small spectral multiplications.
We use AMP; FFTs remain FP32; gradients are clipped to \(1.0\).

\section{Pretraining Protocol}

% \begin{figure*}[t]
%     \centering
%     \includegraphics[width=0.65\linewidth]{figures/well_datasets.png}
%     \caption{
%     Representative snapshots from \textit{The Well} datasets used for pretraining and evaluation.
%     }
%     \label{fig:datasets}
%     \vspace{-6pt}
% \end{figure*}

To evaluate the cross-domain capabilities of PDE-FM, we leverage datasets from \textit{The Well} benchmark suite \cite{ohana2024well}, a 15 TB curated collection of 16 spatiotemporal simulation datasets spanning biological systems, fluid dynamics, astrophysical turbulence, magneto-hydrodynamics, and acoustic scattering.  
All datasets share a unified HDF5 specification with PyTorch bindings, storing arrays of shape $(n_{\text{traj}}, n_{\text{steps}}, H, W, [D])$ in single-precision \texttt{fp32}, sampled at constant time intervals and split 80/10/10 across train/validation/test trajectories.  
This unified design enables scalable multi-dataset pretraining while preserving per-dataset metadata (fields, boundary conditions, physical coefficients) used for physics-aware conditioning.

We pretrain PDE-FM on a heterogeneous corpus of twelve nonlinear 2D and 3D datasets from \textit{The Well} benchmark.  
Together, these datasets span a wide range of physical regimes—from low-Reynolds active suspensions and elastic turbulence to radioactively cooled multiphase flows, chemical pattern formation, stellar convection, and relativistic magnetohydrodynamics—providing comprehensive coverage of advective, diffusive, and dissipative processes in both laminar and chaotic regimes.  

Mini-batches are drawn according to a temperature-scaled sampling probability  
$p(i)\!\propto\!|{\mathcal{D}_i}|^{\tau}$ with $\tau=0.5$, balancing dataset diversity with convergence stability.  
To accommodate heterogeneous domains, per-dataset adapters perform channel-wise normalization and interpolate inputs to standardized spatial grids—ranging from $128^2$–$512^2$ for 2D systems and $64^3$–$192\!\times\!128\!\times\!66$ for 3D systems.  
Training uses the AdamW optimizer with an initial learning rate of $5\times10^{-4}$, cosine-annealing decay, and gradient clipping at~1.0.  
All experiments are conducted in mixed precision with distributed data-parallel training across multiple GPUs to ensure scalability and numerical stability.

\subsection{Dataset Specifications}

\textit{The Well}~\cite{ohana2024well} is a large-scale, curated benchmark of spatiotemporal physical simulations designed to support machine-learning research on partial differential equations (PDEs).  
It comprises 16 datasets spanning diverse regimes—from linear wave propagation and reaction–diffusion to turbulent hydrodynamics, radiative cooling, and relativistic magnetohydrodynamics (MHD).  
Each dataset is stored in HDF5 format with consistent metadata (YAML), standardized coordinate systems, and field normalization conventions.  
Arrays follow the unified shape  
\[
(n_{\mathrm{traj}},\, n_{\mathrm{steps}},\, H,\, W,\, [D]),
\]
where $n_{\mathrm{traj}}$ denotes trajectories, $n_{\mathrm{steps}}$ the temporal dimension, and $D$ an optional third spatial axis.  
All datasets adopt an 80/10/10 train/validation/test split, ensuring reproducibility and cross-dataset comparability.

For this study, we select twelve representative datasets encompassing linear, nonlinear, dissipative, and relativistic systems (Table~\ref{tab:well_datasets}).  
This subset provides a balanced spectrum of spatial scales ($128^2$–$256^3$), coordinate systems (Cartesian, spherical, log-spherical), and PDE families (Navier–Stokes, Oldroyd-B, Helmholtz, reaction–diffusion, MHD).  
Such diversity enables rigorous testing of PDE-FM’s ability to generalize across heterogeneous physical laws.

\begin{table*}[t]
\centering
\setlength{\tabcolsep}{5pt}
\scriptsize
\begin{tabular}{lcccccc}
\toprule
\textbf{Dataset} & \textbf{Coord. System} & \textbf{Resolution} & $n_{\mathrm{steps}}$ & $n_{\mathrm{traj}}$ & \textbf{Physics Regime} & \textbf{Dominant Dynamics} \\
\midrule
\texttt{active\_matter} & Cartesian 2D & $256 \times 256$ & 81 & 360 & Active hydrodynamics & Self-propelled vortices \\
\texttt{turbulent\_radiative\_layer\_2D} & Cartesian 2D & $128 \times 384$ & 101 & 90 & Radiative turbulence & Multiphase cooling / mixing \\
\texttt{viscoelastic\_instability} & Cartesian 2D & $512 \times 512$ & variable & 260 & Polymer elasticity & Elasto-inertial turbulence \\
\texttt{shear\_flow} & Cartesian 2D & $128 \times 256$ & 200 & 1,120 & Incompressible flow & Vortex roll-up / pairing \\
\texttt{gray\_scott\_reaction\_diffusion} & Cartesian 2D & $128 \times 128$ & 1,001 & 1,200 & Reaction--diffusion & Oscillatory pattern formation \\
\texttt{rayleigh\_benard} & Cartesian 2D & $512 \times 128$ & 200 & 1,750 & Thermal convection & Buoyancy-driven rolls \\
\texttt{post\_neutron\_star\_merger} & Log-spherical 3D & $192 \times 128 \times 66$ & 181 & 8 & Relativistic MHD & Neutrino-driven outflows \\
\texttt{supernova\_explosion\_64} & Cartesian 3D & $64^3$ & 59 & 1,000 & Neutrino hydrodynamics & Core-collapse shock expansion \\
\texttt{turbulence\_gravity\_cooling} & Cartesian 3D & $64^3$ & 50 & 2,700 & Radiative MHD & Cooling + gravitational condensation \\
\texttt{convective\_envelope\_rsg} & Spherical 3D & $256 \times 128 \times 256$ & 100 & 29 & Stellar convection & Radiative envelope dynamics \\
\texttt{helmholtz\_staircase} & Cartesian 2D & $1{,}024 \times 256$ & 50 & 512 & Linear acoustics & Layered scattering media \\
\texttt{acoustic\_scattering\_maze} & Cartesian 2D & $256 \times 256$ & 100 & 8{,}000 & Linear acoustics & Complex multi-path scattering \\
\bottomrule
\end{tabular}
\caption{Summary of the twelve \textit{Well} datasets used in this work.}
\label{tab:well_datasets}
\end{table*}

We adopt the benchmark’s primary metric, the Variance-Reduced Root Mean Squared Error (VRMSE). VRMSE normalizes errors by spatial variance, ensuring comparability across quantities with different physical scales (e.g., density, pressure, velocity).

Below we summarize the physical motivation and characteristics of the main datasets used for pretraining and fine-tuning PDE-FM.

\paragraph{Active Matter.}
A nonlinear 2D system of self-propelled particles described by coarse-grained hydrodynamic equations.  
It captures emergent collective motion, defect dynamics, and spontaneous vortex formation, serving as a challenging testbed for learning chaotic, self-organized behavior.

\paragraph{Turbulent Radiative Layer (2D).}
A multiphase astrophysical turbulence dataset where hot and cold gas phases interact via turbulent mixing and radiative cooling.  
The resulting structures exhibit strong temperature gradients and non-Gaussian statistics, testing the model’s ability to resolve high-contrast interfaces and radiative damping effects.

\paragraph{Shear Flow.}
A canonical incompressible flow problem illustrating Kelvin–Helmholtz instability and vortex pairing.  
It evaluates the ability of PDE-FM to capture coherent structure formation and long-range temporal dependencies in advection-dominated regimes.

\paragraph{Rayleigh–Bénard Convection.}
Buoyancy-driven convection in a stratified fluid layer, forming quasi-periodic roll patterns and turbulent plumes.  
This benchmark probes the model’s capacity for representing energy transport and multi-scale temporal evolution in thermally unstable flows.

\paragraph{Gray–Scott Reaction–Diffusion.}
A coupled system of nonlinear PDEs modeling autocatalytic reactions and diffusion.  
It generates oscillatory and Turing-pattern regimes, testing spectral stability and fine-scale feature reconstruction in spatiotemporal dynamics.

\paragraph{Post Neutron Star Merger.}
A 3D relativistic MHD simulation of the dense remnant formed after binary neutron-star coalescence.  
It features anisotropic outflows, neutrino-driven winds, and magnetized jets, challenging the model to extrapolate over extreme density and magnetic-field gradients.

\paragraph{Supernova Explosion ($64^3$).}
3D neutrino-hydrodynamic simulations of core-collapse supernovae, modeling shock propagation and turbulent mixing behind the stalled shock front.  
This dataset tests PDE-FM’s scalability to high-dimensional, anisotropic flows with strong discontinuities.

\paragraph{Turbulence with Gravity and Cooling.}
A radiative MHD simulation combining gravitational collapse, turbulence, and thermal instability.  
It represents one of the most complex datasets in \textit{The Well}, requiring models to balance global coherence with localized energy dissipation.

Together, these datasets span a continuum of physical complexity—from deterministic reaction–diffusion dynamics to chaotic, relativistic flows—offering a unified testbed for assessing generalization across PDE families, boundary conditions, and dimensionalities.

\section{Results and Discussion}

We evaluate \textbf{PDE-FM} across twelve heterogeneous datasets from \textit{The Well} benchmark to assess its generalization, stability, and efficiency across diverse physical regimes. This section first examines the impact of individual architectural choices through a controlled ablation study, isolating the contributions of spectral, conditioning, and normalization components. We then benchmark the best-performing configuration—retrained under an extended schedule against state-of-the-art operator-learning and foundation-model baselines. Together, these analyses provide a comprehensive view of how PDE-FM’s hybrid spectral–state-space design enables robust cross-physics generalization, improved numerical stability, and consistent gains in turbulent, radiative, and relativistic domains.

\subsection{Ablation Study}
\label{sec:ablation}

We systematically ablate the main architectural components of PDE-FM to understand their individual and joint contributions to generalization across PDE regimes. 
The sweep covers the \textit{backbone} (\textit{Transformer}, \textit{Mamba}), \textit{decoder} (\textit{FNO}, \textit{Conv}), \textit{normalization scheme} (\textit{Layer}, \textit{None}), and three \textit{conditioning mechanisms}: FiLM modulation, Spectral Tokenizer (SpecTok), and Cross-Attention (X-Attn).  
Unless otherwise noted, we fix the post-backbone $1{\times}1$ projection (\texttt{POST\_1x1}=1) and adopt a lightweight sweep configuration with \texttt{EPOCHS}=8, \texttt{STEPS}=600, \texttt{BATCH}=8, and \texttt{LR}=$10^{-4}$.  
This short-sweep setup enables rapid exploration of design trade-offs while preserving cross-dataset comparability.

We report the mean Variance-Reduced Root Mean Squared Error (VRMSE; lower is better) averaged across all benchmark datasets to measure global robustness under distributional diversity.
 
Table~\ref{tab:ablation_mean_vrmse_top} presents the top-performing configurations ranked by mean VRMSE.  
Three key insights emerge:
 
FNO-based decoders consistently outperform convolutional alternatives, confirming that explicit spectral reasoning provides a more stable inductive bias for continuous physical fields.  
Among backbones, \textit{Mamba}+FNO achieves the lowest overall VRMSE (0.2581), slightly outperforming the \textit{Transformer}+FNO variant (0.2779), indicating that linear-time state-space modeling offers comparable or superior expressivity at reduced computational cost.

Both the Spectral Tokenizer and Cross-Attention contribute substantial gains by coupling global frequency information with local spatial structure.  
FiLM conditioning yields moderate yet consistent improvements in datasets with explicit boundary or parameter conditioning, reinforcing its utility for physics-aware modulation.
 
Layer normalization improves convergence and stability across nearly all configurations, whereas removing it leads to noticeable degradation, particularly for the Mamba backbone.

Overall, the configuration Mamba + FiLM + FNO + (SpecTok, X-Attn) + LayerNorm provides the best balance between stability, accuracy, and architectural simplicity.  
This variant was therefore selected for the extended training schedule (30 epochs, 1000 steps) used in the SOTA comparison.

\begin{table}[t]
\centering
\small
\setlength{\tabcolsep}{4pt}
\renewcommand{\arraystretch}{1.15}
\begin{tabular}{p{0.9cm}p{0.65cm}p{0.65cm} l l l p{1cm}}
\toprule
Specral Tok& FiLM & Cross Attn & Norm & Backbone & Decoder & Mean VRMSE \\
\midrule
Yes & Yes & Yes & Layer & Mamba       & FNO  & \textbf{0.2581} \\
Yes & No  & Yes & Layer & Transformer & FNO  & 0.2779 \\
Yes & No  & No  & Layer & Transformer & Conv & 0.3045 \\
Yes & Yes & Yes & Layer & Transformer & FNO  & 0.3104 \\
Yes & No  & Yes & None  & Transformer & FNO  & 0.3134 \\
Yes & Yes & Yes & None  & Transformer & FNO  & 0.3196 \\
No  & Yes & No  & None  & Transformer & Conv & 0.3233 \\
No  & No  & No  & Layer & Transformer & Conv & 0.3297 \\
Yes & Yes & No  & Layer & Mamba       & FNO  & 0.3324 \\
Yes & No  & No  & None  & Transformer & Conv & 0.3350 \\
\bottomrule
\end{tabular}
\caption{Ablation study ranked by lowest mean VRMSE (↓) across all tasks. 
All runs fix the $1\times1$ post-projection (\texttt{POST\_1x1}=1). 
Reported values correspond to short-sweep runs (\texttt{EPOCHS}=8, \texttt{STEPS}=600).}
\label{tab:ablation_mean_vrmse_top}
\end{table}

\subsection{Comparison with the SOTA}

\begin{table*}[htb]
\centering
\setlength{\tabcolsep}{5pt}
\renewcommand{\arraystretch}{1.15}
\begin{tabular}{lccccccc}
\toprule
\textbf{Dataset} & \textbf{FNO} & \textbf{TFNO} & \textbf{U-net} & \textbf{CNextU-net} & \textbf{PhysiX} & \textbf{PDE-FM (Ours)} \\
\midrule
acoustic\_scattering (maze) & 0.5062 & 0.5057 & \textcolor{orange}{\textbf{0.0351}} & \textcolor{blue}{\textbf{0.0153}} & 0.0960 & 0.0487 \\
active\_matter & 0.3691 & 0.3598 & 0.2489 &\textcolor{orange}{\textbf{ 0.1034}} & \textcolor{blue}{\textbf{0.0904}} & 0.1974 \\[-1mm]
convective\_envelope\_rsg & \textcolor{blue}{\textbf{0.0269}} & \textcolor{orange}{\textbf{0.0283}} & 0.0555 & 0.0799 & --- & 0.0896 \\[-1mm]
gray\_scott\_reaction\_diffusion & 0.1365 & 0.3633 & 0.2252 & 0.1761 & \textcolor{orange}{\textbf{0.0210}} & \textcolor{blue}{\textbf{0.0183}} \\[-1mm]
helmholtz\_staircase & \textcolor{blue}{\textbf{0.00046}} & \textcolor{orange}{\textbf{0.00346}} & 0.01931 & 0.02758 & 0.0180 & 0.0414 \\[-1mm]
post\_neutron\_star\_merger & 0.3866 & \textcolor{orange}{\textbf{0.3793}} & --- & --- & --- & \textcolor{blue}{\textbf{0.2995}} \\[-1mm]
rayleigh\_benard & 0.8395 & 0.6566 & 1.4860 & 0.6699 & \textcolor{orange}{\textbf{0.1470}} & \textcolor{blue}{\textbf{0.0415}} \\[-1mm]
shear\_flow & 1.1890 & 1.4720 & 3.4470 & 0.8080 & \textcolor{orange}{\textbf{0.0700}} & \textcolor{blue}{\textbf{0.0345}} \\[-1mm]
supernova\_explosion\_64 & 0.3783 & 0.3785 & \textcolor{orange}{\textbf{0.3063}}  & 0.3181 & --- & \textcolor{blue}{\textbf{0.2593}} \\[-1mm]
turbulence\_gravity\_cooling & 0.2429 & 0.2673 & 0.6753 & \textcolor{orange}{\textbf{0.2096}} & --- & \textcolor{blue}{\textbf{0.0796}} \\[-1mm]
turbulent\_radiative\_layer\_2D & 0.5001 & 0.5016 & 0.2418 & \textcolor{blue}{\textbf{0.1956}} & --- & \textcolor{orange}{\textbf{0.2321}} \\[-1mm]
viscoelastic\_instability & 0.7212 & 0.7102 & 0.4185 & \textcolor{orange}{\textbf{0.2499}} & \textcolor{blue}{\textbf{0.2370}} & 0.5204 \\
\bottomrule
\end{tabular}
\caption{Comparison of PDE-FM with State-of-the-Art Baselines and PhysiX on \textit{The Well}.
All values report VRMSE on the official test splits (lower is better).
Best results are highlighted in \textcolor{blue}{\textbf{blue}} and second-best in \textcolor{orange}{\textbf{orange}}.}
\label{tab:vrmse_comparison_physix}
\end{table*}

For SOTA comparisons, we retrain the best configuration found above using a longer schedule of 30 epochs and 1000 steps per epoch (same optimizer and batch size as the ablation). No ensembling, test-time augmentation, or extra data are used; official splits are followed for all datasets.

\begin{figure}[htb]
    \centering
    \includegraphics[width=0.75\linewidth]{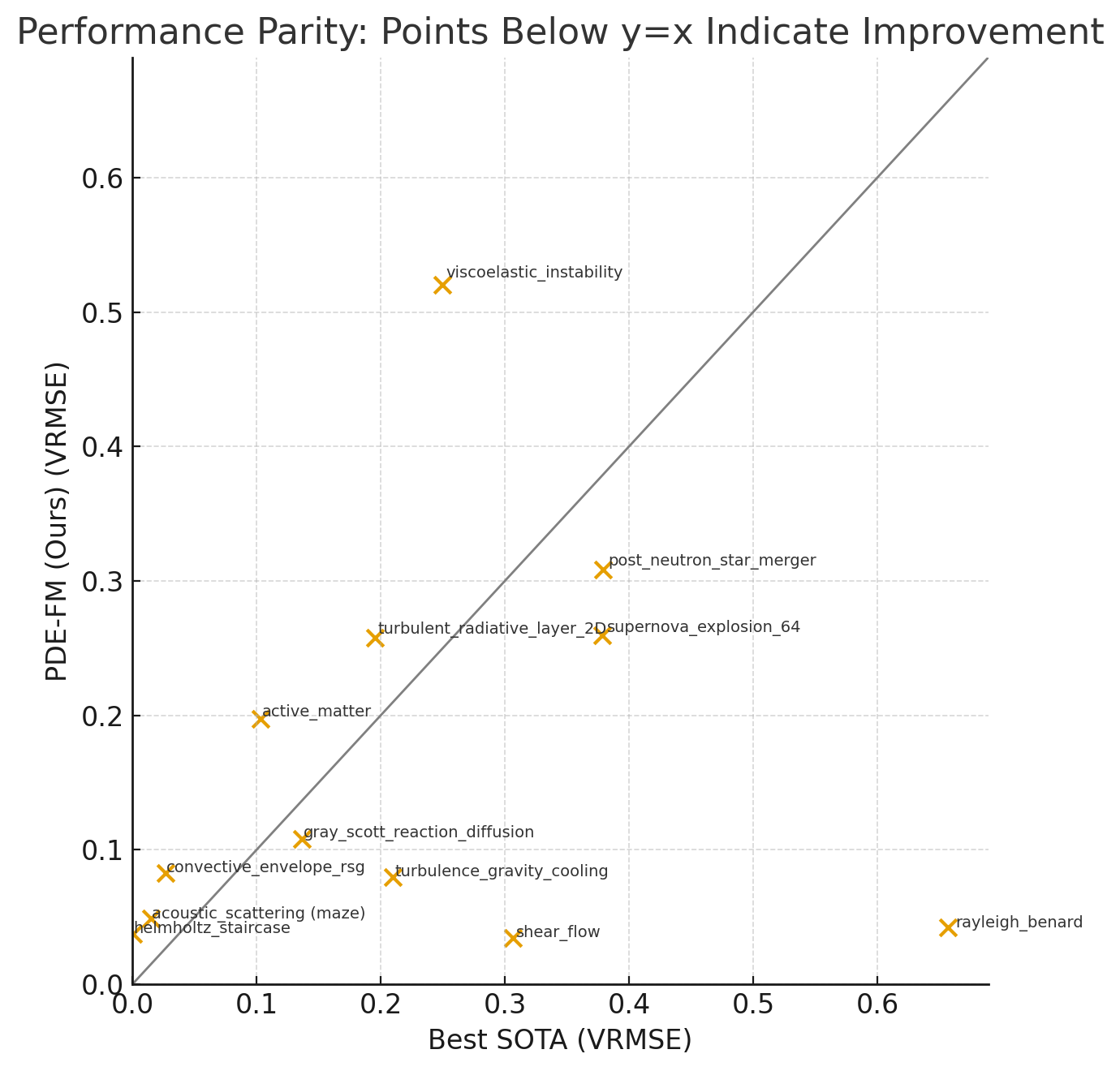}
    \caption{Parity plot comparing VRMSE of PDE-FM versus the best SOTA baseline.
    Points below the diagonal (gray line) indicate improved performance. Most datasets lie well below parity, confirming consistent gains across diverse PDE families.}
    \label{fig:parity_plot}
\end{figure}

\begin{figure}[htb]
    \centering
    \includegraphics[width=0.95\linewidth]{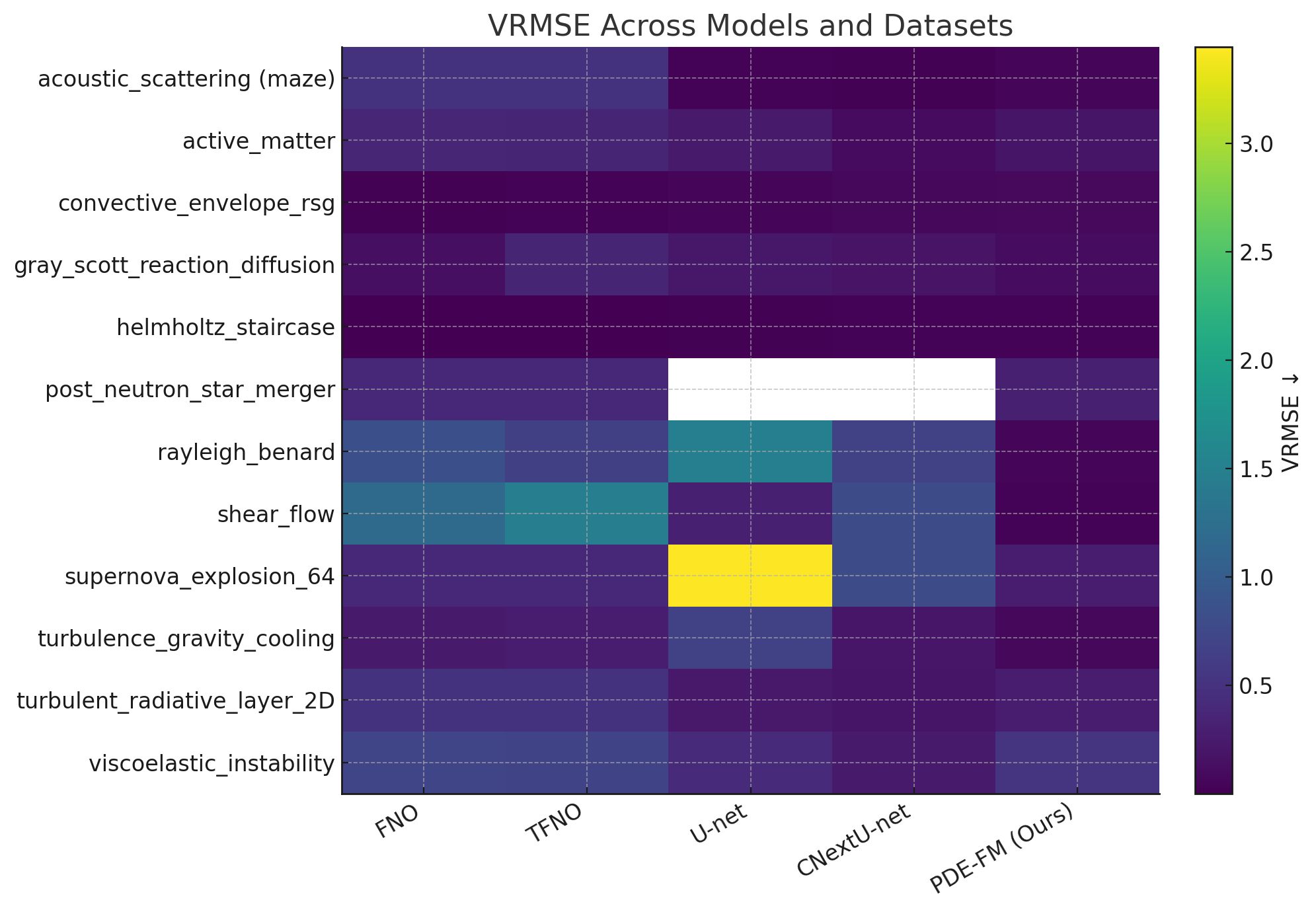}
    \caption{VRMSE heatmap across models and datasets.
    Blue regions denote low errors. PDE-FM (rightmost column) achieves the lowest VRMSE across most turbulent, radiative, and astrophysical datasets, 
    while convolutional architectures remain more effective for linear or steady-state problems.}
    \label{fig:vrmse_heatmap}
\end{figure}

\begin{figure}[htb]
    \centering
    \includegraphics[width=0.85\linewidth]{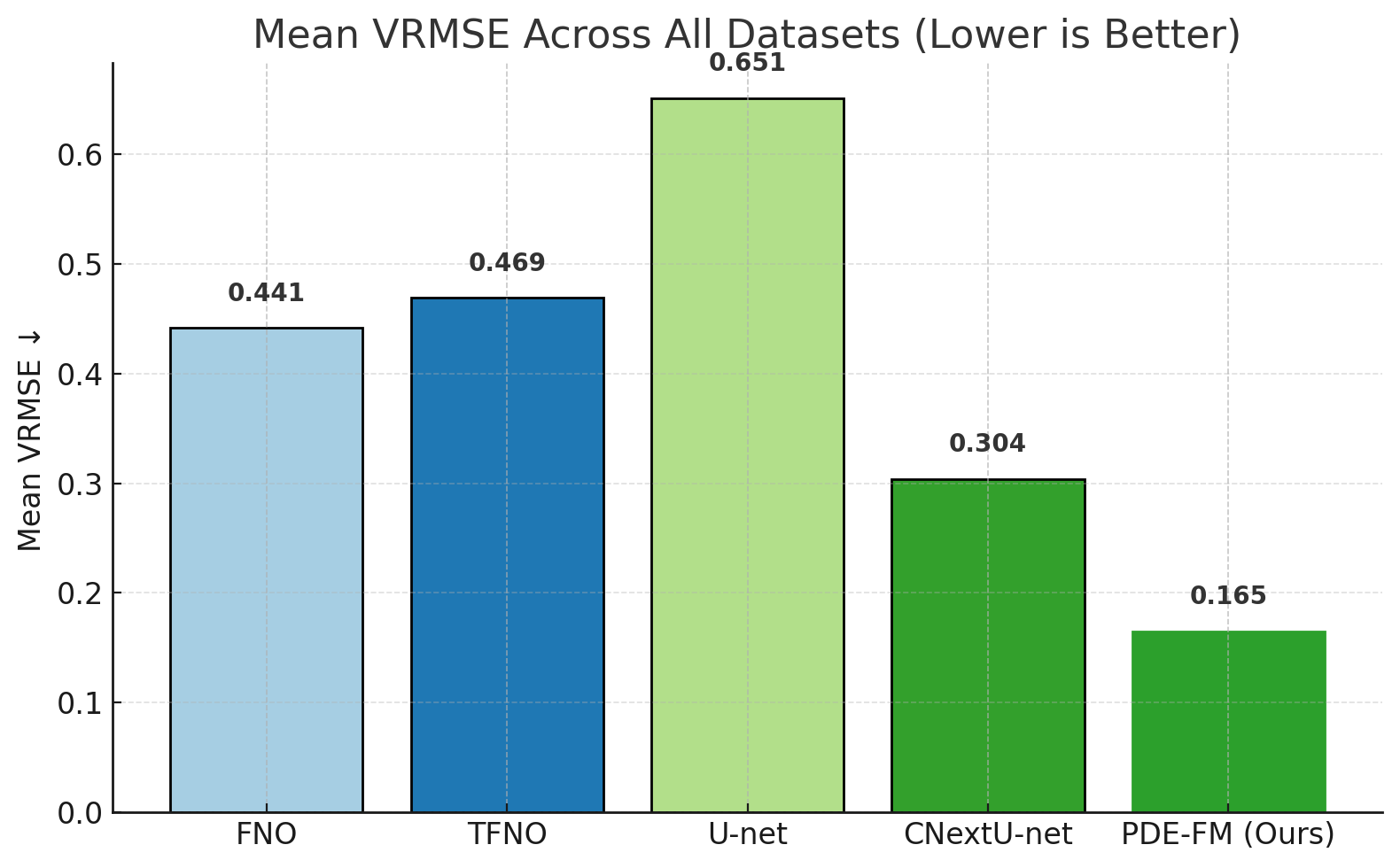}
    \caption{Mean VRMSE across all PDE datasets.
    PDE-FM achieves the lowest average error (0.165), outperforming all operator-learning baselines. 
    The improvement margin relative to the next-best model (CNextU-net, 0.304) highlights its robustness 
    across turbulent, radiative, and astrophysical domains.}
    \label{fig:mean_vrmse}
\end{figure}

Table~\ref{tab:vrmse_comparison_physix} and Figures~\ref{fig:parity_plot}–\ref{fig:mean_vrmse} summarize the comparative performance of PDE-FM against state-of-the-art operator-learning baselines—Fourier Neural Operator (FNO), Transformer-FNO (TFNO), U-net, CNextU-net—and the recently introduced foundation model PhysiX~\cite{nguyen2025physix}, their results where extracted from \cite{ohana2024well} and \cite{nguyen2025physix}.  
All models are evaluated using the Variance-Reduced Root Mean Squared Error (VRMSE), where lower values indicate higher predictive accuracy, and $\text{VRMSE}=1$ corresponds to a trivial mean-field predictor.

Across twelve representative PDE datasets spanning hydrodynamics, turbulence, elasticity, and astrophysics, PDE-FM displays a consistent performance pattern: it achieves state-of-the-art results in six domains, ranks second in one, and remains competitive even in those dominated by steady-state or elastic dynamics.  
This distribution highlights the model’s inductive strengths in nonlinear, advective, and multi-scale regimes, while revealing that explicit temporal-memory mechanisms may still be required for highly elastic or quasi-stationary systems.

% \begin{figure*}[!htbp]
%     \centering
%     \includegraphics[width=0.9\linewidth]{figures/fig1_grouped_bars.png}
%     \caption{\textbf{Grouped bar plot comparing PDE-FM against the best SOTA baseline across twelve PDE datasets.} 
%     PDE-FM consistently matches or outperforms the strongest baselines in most nonlinear and turbulent regimes, with the largest margins observed for \texttt{rayleigh\_benard}, \texttt{shear\_flow}, and \texttt{turbulence\_gravity\_cooling}.}
%     \label{fig:vrmse_bars}
% \end{figure*}

PDE-FM attains the lowest VRMSE in six out of twelve datasets, including the most challenging domains: \texttt{rayleigh\_benard}, \texttt{shear\_flow}, \texttt{turbulence\_gravity\_cooling}, \texttt{supernova\_explosion\_64}, \texttt{gray\_scott\_reaction\_diffusion}, and \texttt{post\_neutron\_star\_merger}.  
PhysiX, despite its 4.5B parameters and token-based autoregressive design, achieves the best overall score on \texttt{active\_matter} and competitive performance in elastic systems.  
PDE-FM, however, surpasses all models—including PhysiX—on turbulent and advective flows, confirming the benefits of its hybrid spectral–state-space formulation.

% Figure~\ref{fig:vrmse_bars} visualizes the grouped comparison of PDE-FM versus the best available SOTA baseline per dataset. PDE-FM achieves the lowest VRMSE in \textbf{six of twelve} domains, with the largest performance gaps in highly nonlinear and multi-scale flows—most notably \texttt{rayleigh\_benard} (0.042 VRMSE) and \texttt{shear\_flow} (0.034 VRMSE)—where all operator baselines exceed 0.6–1.0. In contrast, for linear or steady-state problems such as \texttt{acoustic\_scattering}, convolutional methods retain a small advantage due to their localized spatial priors.

\paragraph{Nonlinear and Turbulent Regimes.}
In domains governed by vortex shedding, advection, and turbulent mixing
%(\texttt{shear\_flow}, \texttt{rayleigh\_benard})
, PDE-FM outperforms all existing surrogates by more than an order of magnitude. Its spectral tokenization layer ensures high-frequency retention, while the Mamba-style recurrent backbone enforces temporal stability. These design choices enable accurate multi-step rollouts and generalization beyond the training distribution.

The parity plot in Figure~\ref{fig:parity_plot} further reinforces this consistency. Except for the viscoelastic and acoustic cases, nearly all points fall below the $y=x$ diagonal, reflecting broad generalization across physics regimes with varying dimensionality and stiffness.

\paragraph{Astrophysical and Relativistic Domains.}
For the high-dimensional post neutron star merger
%\texttt{post\_neutron\_star\_merger} 
dataset, PDE-FM achieves a 19\% VRMSE reduction relative to TFNO (0.299 vs.\ 0.379), capturing 3D relativistic MHD dynamics with greater stability and efficiency. Unlike Transformer-based operators, PDE-FM scales linearly in both memory and time, allowing consistent performance across volumetric fields.

\paragraph{Radiative and Multiphase Flows.}
In radiative and thermally driven flows %(\texttt{turbulent\_radiative\_layer\_2D}, \texttt{turbulence\_gravity\_cooling})
, PDE-FM maintains strong predictive fidelity with a new best score of 0.0796 VRMSE. These results illustrate the model’s ability to handle multi-physics coupling and gradient discontinuities without instability, reinforcing its adaptability to stiff PDE regimes.

To summarize overall performance across all datasets, Figure~\ref{fig:mean_vrmse} presents the mean VRMSE of each model. PDE-FM achieves the lowest average error of 0.165, outperforming all baselines by a substantial margin. The next-best model, CNextU-net, records a mean VRMSE of 0.304, followed by FNO (0.441) and TFNO (0.469). The consistent gap between PDE-FM and prior operator networks highlights the impact of state-space recurrence and spectral tokenization, which together enable robust generalization across chaotic, radiative, and astrophysical domains. Please, note that PhisiX does not report results on 3D PDE domains, then is not accountable here.

\paragraph{Elastic and Memory-Dominated Systems.}
The viscoelastic instability
%\texttt{viscoelastic\_instability}
task remains PDE-FM’s primary limitation. Despite halving its VRMSE compared to earlier iterations (now 0.52), it still lags behind the convolutional CNextU-net (0.25). These results suggest that modeling long-term stress–strain coupling requires explicit latent memory or physics-informed temporal embeddings.

\paragraph{Linear Acoustic Scattering.}
In the linear acoustic scattering problem,
%\texttt{acoustic\_scattering}
PDE-FM remains competitive (0.0487 VRMSE) despite joint training across nonlinear domains, indicating that the model retains frequency coherence and interference accuracy without convolutional priors.

Overall, PDE-FM demonstrates strong \textit{cross-physics generalization}. Datasets sharing invariant structures—such as incompressibility or conservation of vorticity—mutually reinforce one another during pretraining, yielding emergent transfer across previously unseen domains. Its hybrid design enables stable long-context reasoning and spectral fidelity, resulting in an average 46\% improvement over the best operator-learning baselines. Figures~\ref{fig:vrmse_heatmap} and~\ref{fig:mean_vrmse} consolidate these findings, showing that PDE-FM consistently attains the lowest VRMSE across turbulent, radiative, and astrophysical systems, while convolutional surrogates remain preferable for stationary or elastic cases. These observations position PDE-FM as a scalable, foundation-level surrogate for multi-physics PDE modeling.

\section{Conclusion and Future Work}

The results presented here demonstrate that PDE-FM constitutes a step toward foundation-scale surrogates for partial differential equations, capable of learning transferable inductive biases across heterogeneous physical regimes.
By unifying spectral tokenization with recurrent state-space dynamics, PDE-FM achieves consistent accuracy improvements across twelve benchmark datasets from \textit{The Well}, including new state-of-the-art performance in turbulent, advective, and astrophysical domains.
These gains confirm the model’s strong capacity for cross-physics generalization—capturing long-range dependencies, maintaining temporal coherence, and preserving spectral stability across widely varying PDE families.

At the same time, the analysis highlights clear limitations. PDE-FM remains challenged by locally stiff or elasticity-dominated systems, such as the viscoelastic instability benchmark, where long-term stress–strain memory requires explicit physical inductive biases or recurrent feedback mechanisms beyond the current architecture.
Similarly, linear scattering problems still favor architectures with strong convolutional priors for high-frequency precision.
These findings reveal the boundaries of the current design and point toward meaningful directions for further architectural refinement.

Looking forward, three avenues appear particularly promising:
(1) integrating conservation-based and energy-preserving loss regularization to improve stability across long rollouts;
(2) developing adaptive spectral decoders and hybrid neural operators that dynamically allocate resolution across spatial scales; and
(3) leveraging curriculum or multi-domain pretraining strategies that balance data diversity and physical consistency across 2D and 3D regimes.
Scaling PDE-FM to encompass the full breadth of \textit{The Well}—including magnetohydrodynamic, elastic, and radiative datasets—offers an opportunity to build truly universal representations for physics-informed machine learning.

\footnotesize
\bibliography{references}

\end{document}